\documentclass[journal,twoside,web]{ieeecolor}
\usepackage[utf8]{inputenc}
\usepackage{tmi}
\usepackage{cite}
\usepackage{amsmath,amssymb,amsfonts}
\usepackage{comment}
\usepackage{algorithmic}
\usepackage{graphicx}
\usepackage{textcomp}
\usepackage{pdfpages}
\usepackage{multicol}
\usepackage{booktabs}
\usepackage{multirow}
\usepackage{bigstrut}
\usepackage{xspace}
\usepackage[table,xcdraw]{xcolor}
\usepackage{adjustbox}
\usepackage{algorithmic,algorithm}
\begin{document}
\title{
UPL-SFDA: Uncertainty-aware Pseudo Label Guided Source-Free Domain Adaptation for Medical Image Segmentation}
\author{Jianghao Wu, Guotai Wang, Ran Gu, Tao Lu, Yinan Chen, Wentao Zhu, Tom Vercauteren, Sébastien~Ourselin, Shaoting~Zhang
\thanks{This work was supported by National Natural Science Foundation of China (No. 62271115) and core funding from the Wellcome/EPSRC [WT203148/Z/16/Z; NS/A000049/1]. TV is supported by a Medtronic/RAEng Research Chair [RCSRF1819/7/34].}
\thanks{Jianghao Wu, Guotai Wang, Ran Gu and Shaoting Zhang are with the School of Mechanical and Electrical Engineering, University of Electronic Science and Technology of China, Chengdu, 611731, China. Guotai Wang and and Shaoting Zhang are also with Shanghai AI laboratory, Shanghai, 200030, China. (e-mail: guotai.wang@uestc.edu.cn, zhangshaoting@uestc.edu.cn)}
\thanks{Tao Lu is with the Department of Radiology, Sichuan Provincial People’s Hospital, University of Electronic Science and Technology of China, Chengdu, 610072, China}
\thanks{Yinan Chen is with SenseTime Research, Shanghai, 200233, China}
\thanks{Wentao Zhu is with Research Center for Healthcare Data Science, Zhejiang Laboratory, Hangzhou, 311100, China}
\thanks{Tom Vercauteren and Sébastien~Ourselin are with the School of Biomedical Engineering
\& Imaging Sciences, King’s College London, London, WC2R 2LS, UK.}
}
\maketitle

\begin{abstract}
\textcolor{black}{Domain Adaptation (DA) is important for deep learning medical image segmentation models to deal with testing images from a new target domain. As the source-domain data are usually unavailable when a trained model is deployed at a new center, Source-Free Domain Adaptation (SFDA) is appealing for data and annotation-efficient adaptation to the target domain. However, existing SFDA methods have a limited performance due to lack of sufficient supervision with source-domain images unavailable and target-domain images unlabeled.} We propose a novel Uncertainty-aware Pseudo Label guided  (UPL) \textcolor{black}{SFDA} method for medical image segmentation. \textcolor{black}{Specifically, we propose \textcolor{black}{Target Domain Growing (TDG)} to enhance the diversity of predictions in the target domain by duplicating the pre-trained model's prediction head multiple times with perturbations.} The different predictions in these duplicated heads are used to obtain pseudo labels for unlabeled target-domain images and their uncertainty to identify reliable pseudo labels. We also propose a \textcolor{black}{Twice Forward pass Supervision (TFS)} strategy that uses reliable pseudo labels obtained in one forward pass to supervise predictions in the next forward pass. The adaptation is further regularized by a mean prediction-based entropy minimization term that encourages confident and consistent results in different prediction heads.   
\textcolor{black}{UPL-SFDA was validated with a multi-site heart MRI segmentation dataset, a cross-modality fetal brain segmentation dataset, and a 3D fetal tissue segmentation dataset. It improved the average Dice by 5.54, 5.01 and 6.89 percentage points for the three tasks compared with the baseline, respectively, and outperformed several state-of-the-art SFDA methods.}  
\end{abstract}


\section{Introduction}
\label{sec:introduction}

\IEEEPARstart {D}{eep} learning has achieved excellent performance in medical image segmentation tasks in recent years~\cite{litjens2017survey,duan2020sensecare}. Its current success is highly dependent on the assumption that training and testing images are from the same distribution. 
However, in practice, a model trained with images from one certain source domain may be used to deal with images in an unseen target domain with different image appearances, which is usually caused by different scanning devices,  imaging protocols, patient groups or image qualities, etc. Failing to deal with the gap between the source and target domains will lead to a dramatic performance decrease~\cite{gu2021domain}. As it is impossible to collect images from all the potential target domains  during training, it is essential to make the model adapted to images in the unseen target domain after deployment. 



Domain Adaptation (DA) that aims to solve the domain gap between training and testing data is attracting increasing attentions recently~\cite{guan2021domain}. 
Though collecting a set of annotated images in the target domain to fine-tune the pre-trained model can make it adapted to the target domain, the annotations are expensive to obtain and usually unavailable in the target domain for model deployment. Therefore, many researchers have investigated  Unsupervised Domain Adaptation (UDA)~\cite{guan2021domain} %
that uses unannotated images in the target domain for adaptation. Most existing UDA methods require access to source-domain and target-domain images simultaneously for training~\cite{pei2021disentangle,wu2022fpl}. 
\textcolor{black}{However, due to concerns on privacy, bandwidth and other issues, it is not always possible to access source-domain data and target-domain data simultaneously.}


\textcolor{black}{Source-Free Domain Adaptation (SFDA)~\cite{wen2023source,li2021free, sun2020test} 
aims to adapt a model pre-trained with source-domain images to fit the target data distribution without access to the source data. 
Due to the absence of annotations in the target domain, the main challenge for SFDA is the lack of sufficient supervision for the model in the target domain.} To deal with this problem, some existing works designed auxiliary tasks such as rotation prediction~\cite{sun2020test}, image normalization~\cite{karani2021test} and auto-encoder-based image reconstruction~\cite{he2021autoencoder} to assist adaptation in the target domain. However, these works introduce an extra sub-network for the auxiliary task that needs to be trained in the  source domain in advance, which makes these \textcolor{black}{SFDA} methods only work for a model pre-trained in a specified way in the source domain and cannot be applied to models pre-trained in other manners, e.g., standard supervised learning without auxiliary tasks. 

\textcolor{black}{In this work, we explore a more flexible approach for SFDA, where only a pre-trained segmentation model and unannotated images are available in the target domain, without restrictions on how the model has been pre-trained in the source domain, and we call it fully SFDA. Note that fully SFDA is independent of the pre-training process, and is more general than the auxiliary task-based methods~\cite{sun2020test,karani2021test,he2021autoencoder} that require special pre-training strategies and network structures. }

To deal with unannotated images in the target domain for fully SFDA, several researchers have investigated some regularization methods, such as  entropy minimization for the predictions in the target domain~\cite{wang2021tent,niu2022efficient}, which are inspired by entropy minimization in the UDA~\cite{li2021attent,liu2021adapting,liu2022self}  and semi-supervised learning tasks\cite{WANG2023pymic,
shi2021inconsistency,LUO2022semi}. 
However, only using entropy minimization as supervision cannot provide sufficient constraints, which makes the model tend to give high-confidence but incorrect predictions in the target domain. To deal with this problem, some researchers also proposed self-training, which fine-tunes the pre-trained model using its predictions on the target-domain images as pseudo labels~\cite{lee2013pseudo,fleuret2021uncertainty,Kingetsu2022}. 
However, due to the change in the target domain distribution, it is hard to obtain accurate pseudo labels, which brings challenges to achieving good performance~\cite{TAJBAKHSH2020101693}.

\textcolor{black}{To overcome these problems, we propose a novel Uncertainty-aware Pseudo Label
guided Source-Free Domain Adaptation (UPL-SFDA) framework for medical image segmentation.} Differently from many existing methods that require a special pre-training strategy in the source domain~\cite{sun2020test,karani2021test,he2021autoencoder}, our method is agnostic to the training stage and has a minimal requirement on the network structure, which is applicable in wider scenarios. Given a pre-trained network, we propose \textcolor{black}{Target Domain Growing (TDG)} that duplicates the prediction head $K$ times in the target domain, and add random perturbations (e.g., dropout, spatial transformation) to obtain $K$ different segmentation predictions. The ensemble of these predictions leads to more robust pseudo labels with efficient uncertainty estimation, which helps to distinguish reliable pseudo labels from unreliable ones. To avoid model degradation commonly faced by self-training, we introduce \textcolor{black}{Twice Forward pass Supervision (TFS)} that uses reliable pseudo labels obtained in one forward pass to supervise predictions in a following forward pass. In addition, unlike existing works imposing entropy minimization on each single prediction head~\cite{wang2021tent,fleuret2021uncertainty}, we impose entropy minimization on the mean prediction across the $K$ heads instead, which  additionally introduces an implicit multi-head consistency regularization to obtain more robust results. Our contributions are summarized as follows:

\begin{itemize}
	\item We propose a  Source-Free Domain Adaptation method based on uncertainty-aware pseudo labels  for medical image segmentation, which adapts a model to the target domain without specific requirements on the pre-training strategy and network structure in the source domain.
	\item We introduce  \textcolor{black}{Target Domain Growing (TDG)} to expand a pre-trained model with perturbed multiple prediction heads in the target domain, which increases the quality of pseudo labels and obtains  uncertainty estimation efficiently. 
	\item A \textcolor{black}{Twice Forward pass Supervision (TFS)} is introduced for self-training, which is combined with a mean prediction-based entropy minimization to robustly learn from pseudo labels in \textcolor{black}{SFDA}. 
    
\end{itemize}
\textcolor{black}{Extensive experiments on three applications (multi-site heart MRI segmentation, cross-modality fetal brain segmentation, and fetal tissue segmentation) showed that our method can effectively adapt the model from a source domain to one or multiple target domains.} It outperformed several existing \textcolor{black}{SFDA} methods for medical image segmentation, and was comparable and even better than supervised training in the target domain. 

\begin{figure*}
    \centering
    \centerline{\includegraphics[width=18cm]{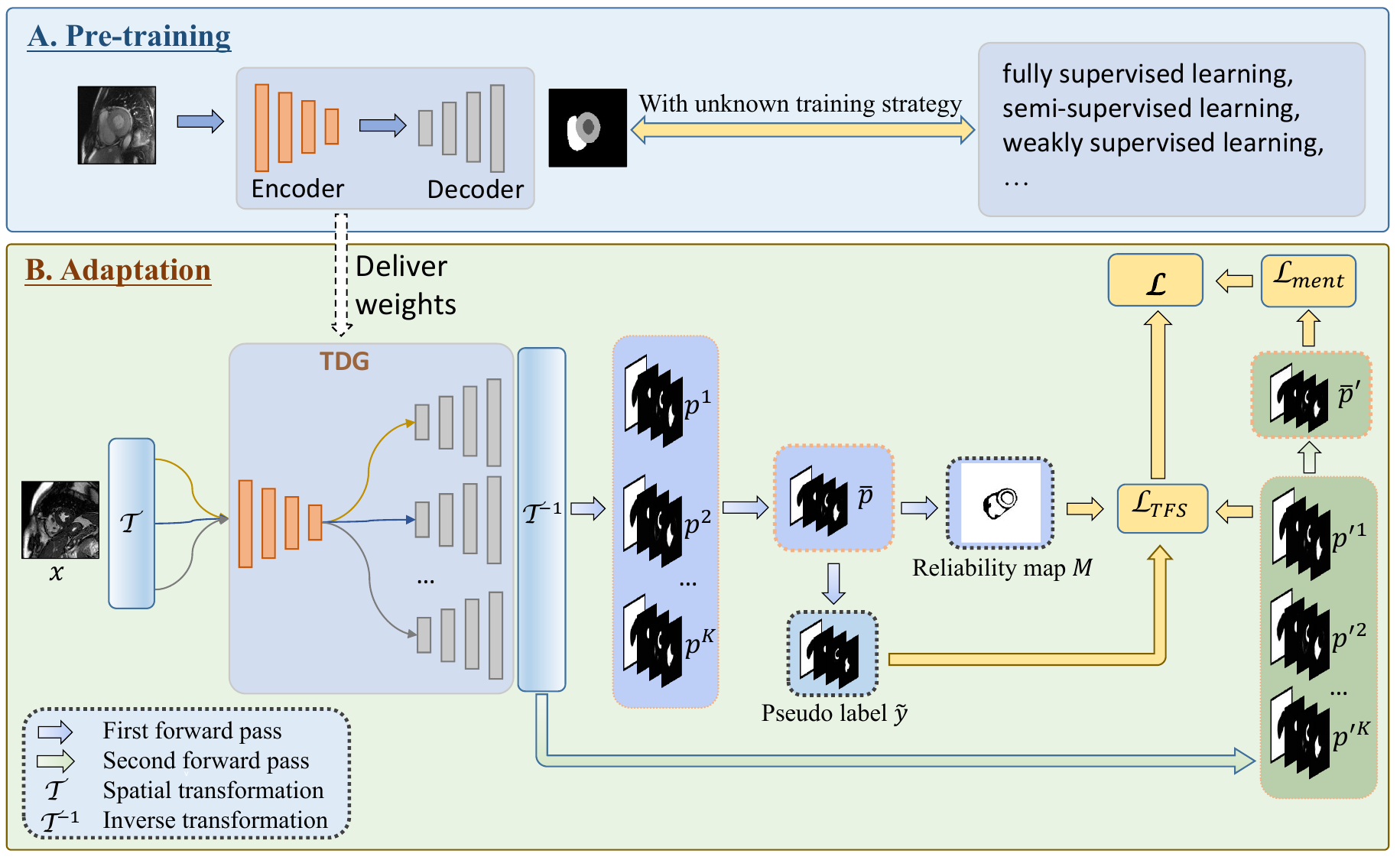}}
    \caption{Overview  of our proposed Uncertainty-aware Pseudo Label guided \textcolor{black}{Source-Free Domain Adaptation (UPL-SFDA)} framework. In the pre-training stage, the model can be trained in the source domain with an arbitrary strategy. We 
    use \textcolor{black}{Target Domain Growing (TDG)} to extend the pre-trained model with multiple prediction heads with perturbations in the target domain. Note that the pseudo label and reliability map obtained in one forward pass are used to supervise the predictions in the next forward pass in the \textcolor{black}{Twice Forward pass Supervision (TFS)} loss. }
    \label{fig:method}
\end{figure*}

\section{Related Works}
\subsection{Unsupervised Domain Adaption}
UDA aims to transfer the knowledge learned from labeled source-domain data to an unlabeled target domain. Current UDA methods mainly adapt the model to the target domain in three aspects. The first is image appearance alignment that translates a target-domain image into a source-domain-like image~\cite{tzeng2017adversarial,dorent2023crossmoda,xu2023novel,wu2023tiss}, 
so that the domain gap is alleviated. The second is feature alignment that minimizes the distance of feature distribution between the source and target domains to learn domain-invariant representations~\cite{chen2020unsupervised}. For example, for cardiac image segmentation, Wu et.~\cite{wu2021unsupervised} used Variational Auto-Encoders (VAEs) to align the features in the source and target domains, and Chen et al.~\cite{chen2019synergistic} used Generative Adversarial networks (GANs) to align the features. The third is output alignment, i.e., using the source model to generate pseudo labels in the target domain for adaptation~\cite{wu2022fpl}. However, even relying on unpaired and unsupervised domain translation techniques, these UDA methods require access to source domain images, which is hardly guaranteed at a testing site due to the concerns on privacy, computational cost and bandwidth. Therefore, source-free DA is highly desirable in practice. 

\subsection{Source-Free Domain Adaption}
\textcolor{black}{Source-Free Domain Adaption (SFDA) 
deals with domain adaption without access to source-domain data~\cite{sun2020test,fleuret2021uncertainty,yang2022source,wen2023source}. 
Yang et al.~\cite{yang2022source} proposed a Fourier-style mining-guided framework, which comprises a generation stage and an adaptation stage for adapting the source model to the target domain using paired source-like and target images.}
Sun et al.~\cite{sun2020test} introduced an auxiliary branch to predict the rotation angle in the target domain.
Karani et al.~\cite{karani2021test} introduced a shallow image normalization network before the segmentation model, and fine-tuned the normalization network in the target domain based on predictions refined by a Denoising Auto-Encoder (DAE). 
However, these methods require the segmentation model's structure to be modified in advance to support the auxiliary task and pre-trained with a specified  strategy, which is inapplicable to general segmentation models that are unaware of the adaptation process during pre-training. \textcolor{black}{Recently, some methods~\cite{nado2020evaluating,wang2021tent} avoid the coupling between training in the source and target domains, so that the adaptation process does not set a prerequisite for training methods in the source domain, which is more general to arbitrary pre-trained models. Wen et al.~\cite{wen2023source} proposed a selectively updated Mean Teacher for SFDA, where predictions from a teacher model based on exponential moving average is used to supervise the student.}  Nado et al.~\cite{nado2020evaluating} proposed Prediction-Time Batch Normalization (PTBN) that recalculates statistics of batch normalization layers according to the images in the target domain. TENT~\cite{wang2021tent} updates the parameters in  batch normalization layers to minimize the entropy of predictions in the target domain. 
In addition to entropy minimization~\cite{wang2021tent}, other loss functions, such as 
regional nuclear-norm loss with contour regularization~\cite{hu2021fully} and consistency regularization~\cite{Thomas2021miccai}, have been proposed for the setting. However, due to the lack of annotations, achieving good performance for \textcolor{black}{SFDA} methods is still challenging.


\section{Method}\label{method}
Fig.~\ref{fig:method} shows an overview of our proposed Uncertainty-aware Pseudo Label guided \textcolor{black}{Source-Free Domain Adaptation (UPL-SFDA)}.
It is independent of the pre-training stage in the source domain, so it can deal with a model 
pre-trained in an arbitrary strategy. 
In \textcolor{black}{UPL-SFDA}, we introduce \textcolor{black}{Target Domain Growing (TDG)} to extend the source model into a multi-head prediction structure by duplicating the pre-trained prediction head $K$ times, and then get pseudo labels based on an ensemble of the prediction heads with perturbations using dropout and spatial transformation. Pseudo labels obtained in one forward pass are used to supervise the prediction of the next forward pass, which acts as a consistency regularization between the two forward passes, and they are weighted by the reliability (confidence).    
For unreliable pixels, we use a mean prediction-based entropy minimization regularization that improves confidence of the predictions and inter-head consistency. 

\subsection{Pre-trained Model from the Source Domain}\label{sec:method_model}
Let $S$ and $T$ be the source  and  target domains, respectively. Let $\mathbf{X}_S$ = $\lbrace (\boldsymbol{x}_i^s, y_i^s) ,i = 1, ..., N_s \rbrace$ 
be the training images and their labels in the source domain, and $\mathbf{X}_T$ = $\lbrace (\boldsymbol{x}_i^t, ), i = 1, ..., N_t \rbrace$ represent unlabeled images in the target domain for adaptation, where $N_s $ and $N_t$ are the number of samples in the two domains, respectively. Note that the data distributions in $S$ and $T$ are different, and we assume that the label has the same distribution across the two domains, i.e., the same type of structure for segmentation.

For a general CNN-based segmentation model, it has a feature extractor $g$ and a prediction head $h$, and the parameters of the segmentation model are denoted as $\{\theta_g, \theta_h\}$, where $\theta_g$ and  $\theta_h$ denote the parameters of $g$ and $h$, respectively. As encoder-decoder networks are widely used for medical image segmentation~\cite{ronneberger2015u, Fabian2021}, we consider $g$ as an encoder and $h$ as a decoder in this work, respectively. 
The model is pre-trained  in the source domain via:

\begin{equation}
\theta_g^0, \theta_h^0  = \text{arg} \mathop{\text{min}}\limits_{\theta_g, \theta_h} \frac{1}{N_s}\sum_{i=1}^{N_s}L_{s}\Big( h\big(g(\boldsymbol{x}_i^s)\big), y_i^s\Big)
\end{equation}
where 
$L_{s}$ donates a certain type of supervision loss in the source domain, which might be implemented by fully supervised learning, semi-supervised learning and weakly supervised learning, etc., based on the type of the available labels in the source domain. $\theta_g^0$ and $\theta_h^0$ denote the optimized parameter values in the source domain, and they are used as initial parameters for the adaptation process in the target domain.  

\subsection{Target Domain Growing in the Target Domain}
With the pre-trained  feature extractor $g$ and prediction head $h$, the model can be applied to a target-domain image to obtain a prediction as the pseudo label. However, due to the gap between source and target domains, directly applying  the pre-trained model will lead to a very low quality of pseudo labels. To improve the quality of pseudo labels for a higher adaptation performance, we propose \textcolor{black}{Target Domain Growing (TDG)} to extend the source model, i.e., we duplicate the prediction head (i.e., decoder) $h$ by $K$ times in the target domain, and they are
initialized as the pre-trained prediction head with parameter values of $\theta^0_h$. These prediction heads are 
connected to the same pre-trained feature extractor $g$ in parallel, as shown in Fig.~\ref{fig:method}. 

Let $h^k$ denote the $k$-th prediction head in the target domain. As they have the same initial parameter values with the same architecture, their outputs will be the same for a given input. To obtain diversity, we introduce perturbations for the prediction heads so that they produce different results for more robust ensemble. Specifically, we use random spatial transformation and dropout to improve the diversity of predictions. 

First, for an input image $\boldsymbol{x} \in \mathcal{R}^{H\times W}$ in the target domain, where $H$ and $W$ are the height and width, respectively, we send it into the network $K$ times, each time with a random spatial transformation and for a different prediction head $h^k$. The segmentation prediction result for the $k$-th head is:
\begin{equation}
    \boldsymbol{p}^k = \mathcal{T}^{-1}_k \circ h^k\big(g(\mathcal{T}_k \circ x)\big)
\end{equation}
where $\mathcal{T}_k$ is a random spatial transformation and $\mathcal{T}^{-1}_k$ is the corresponding inverse transformation.  $\boldsymbol{p}^k \in \mathcal{R}^{C \times  H\times W}$ is the output segmentation probability map with $C$ channels obtained by Softmax, where $C$ is the class number for segmentation. In this paper, we set $\mathcal{T}_k$ as random flipping, random rotation with $\pi/2$, $\pi$ and $3\pi/2$ for efficient implementation. 

Second, we add a dropout layer before each of the prediction head $h^k$, so that the prediction heads take different random subsets of the features as input. Due to the image-level and feature-level perturbations, the $K$ predictions are different for an input image. We then average across the $K$ predicted segmentation probability maps for ensemble:

\begin{equation}
    \bar{\boldsymbol{p}} = \frac{1}{K}\sum_{k=1}^K \boldsymbol{p}^k
\end{equation}

\subsection{Twice forward pass supervision with Reliable Pseudo Labels}\label{sec:method_pseudo_label}
With the average probability prediction $\bar{\boldsymbol{p}}$, we take an argmax to obtain the pseudo label for the input $\boldsymbol{x}$. To reduce noises, we post-process it by only keeping the largest component for each foreground class in segmentation tasks where each foreground class has only one component (e.g, heart structure  and fetal brain segmentation in this work). Then the post-processed pseudo label is converted into a one-hot representation, which is denoted as $\tilde{y} \in \{0, 1\}^{C\times H\times W}$. 

As the domain gap may limit the quality of the pseudo label $\tilde{y}$, directly using $\tilde{y}$ to supervise the network will lead to a limited performance. To deal with this problem, we use  the uncertainty information in $\bar{\boldsymbol{p}}$ to identify pixels with reliable pseudo labels and only use the reliable region to supervise the network.  To achieve this, we define a binary reliability map $M \in \{0, 1\}^{H\times W}$ for $\tilde{y}$, and each element in $M$ is defined as:
\begin{equation}
M_{n} = 
\begin{cases}
1  &  \text{if~$\bar{\boldsymbol{p}}_{c^*,n}$ $>$ $\tau$ }  \\
0  &  \text{otherwise}	\end{cases}	
\end{equation}
where $n$ = 1, 2, ..., $HW$ is the pixel index. $c^*$ = $\arg\max_{c} (\bar{\boldsymbol{p}}_{c,n})$ is the class with the highest probability for pixel $n$, and $\bar{\boldsymbol{p}}_{c^*,n}$ represents the confidence for the pseudo label at that pixel. $\tau \in (1/C, 1.0)$ is a confidence threshold.
For pseudo label-based self-training, the model may be biased towards its own prediction in each iteration. To avoid this problem, Chen et al.~\cite{Chen2021cps} introduced  cross supervision where two networks with different predictions guide each other to reduce the bias. However, the use of two networks would increase the computational and memory cost, and it is not suitable for \textcolor{black}{SFDA} where only one pre-trained model is provided. Inspired by Chen et al.~\cite{Chen2021cps} and to improve the robustness of pseudo label-based \textcolor{black}{SFDA}, we introduce \textcolor{black}{Twice Forward pass Supervision (TFS)} for robust adaptation. 

\textcolor{black}{Specifically, for a batch of data in the training set, before each gradient back-propagation, we perform two consecutive forward passes. We employ the pseudo label $\tilde{y}$ and its associated reliability map $M$ obtained in the first forward pass to supervise the prediction heads in the second forward pass}. Let $\boldsymbol{p}'^k$ denote the output of the $k$-th prediction head in the second forward pass. Due to the use of random spatial transformation and dropout as mentioned above, the outputs of the two forward passes are different despite the same parameter values. Using  $\tilde{y}$ to supervise $\boldsymbol{p}'^k$ can introduce a consistency regularization under perturbations, which improves the robustness of the network. The \textcolor{black}{TFS} loss is:
\begin{table*}
  \centering
  \caption{Details of datasets used for experiments. The values represent volume numbers. 
}
    \begin{adjustbox}{width=0.65\textwidth}
    \begin{tabular}{c|cccc|cc|cc}
    \hline
    Dataset & \multicolumn{4}{c|}{M\&MS Dataset } & \multicolumn{2}{c|}{FB Dataset} & \multicolumn{2}{c}{\textcolor{black}{FeTA Dataset}} \bigstrut\\
    \hline
    \multirow{2}[4]{*}{Domain} & A     & B     & C     & D     & Source & Target & \textcolor{black}{Source} & \textcolor{black}{Target} \bigstrut\\
\cline{2-9}          & Siemens  & Philips  & General Electric    &  Canon & HASTE  & trueFISP &\textcolor{black}{IRTK} &\textcolor{black}{mialSR} \bigstrut\\
    \hline
    Training & 135   & 177   & 105   & 70    & 47    & 30    & \textcolor{black}{28}    & \textcolor{black}{28} \bigstrut[t]\\
    Validation & 19    & 25    & 15    & 10    & 7     & 5     & \textcolor{black}{4}     & \textcolor{black}{4} \\
    Testing & 38    & 50    & 30    & 20    & 14    & 9     & \textcolor{black}{8}     & \textcolor{black}{8} \bigstrut[b]\\
    \hline
    Overall & 192   & 252   & 150   & 100   & 68    & 44    & \textcolor{black}{40}    & \textcolor{black}{40} \bigstrut\\
    \hline
    \end{tabular}%
    \end{adjustbox}
  \label{tab:dataset}
\end{table*}%
\begin{equation}\label{eq_lcp}
\textcolor{black}{\mathcal{L}_{TFS}} = \frac{1}{K}\sum_{k = 1}^K \mathcal{L}_{w-dice}(\boldsymbol{p}'^k, \tilde{y}, M)
\end{equation}
where $\mathcal{L}_{w-dice}$ is the reliability map-weighted Dice loss for a single head. \textcolor{black}{Here we use a Dice-based loss for  pseudo label supervision, as Dice loss can better deal with class imbalance in segmentation tasks than cross entropy~\cite{Milletari2016}, and the segmentation performance is usually evaluated by Dice.}
\begin{equation}
    \mathcal{L}_{w-dice}
    = 1-\frac{1}{C}\sum_{c=1}^C\frac{\sum_{n} 2M_{n}\boldsymbol{p}'^k_{c, n}\tilde{y}_{c,n}}{ \sum_{n}M_n(\boldsymbol{p}'^k_{c, n} + \tilde{y}_{c,n}) +\eta}
\end{equation}
where $n$ is the pixel index and $\eta = 10^{-5}$ is a small number for numeric stability. 

\subsection{Mean Prediction-based Entropy Minimization}
Entropy minimization is widely used for regularization in semi-supervised learning~\cite{LUO2022semi} and SFDA~\cite{lee2022confidence,niu2022efficient,tomar2021opttta}, which improves the model's confidence by minimizing the entropy of the class distribution in a prediction output. 
However, existing entropy minimization methods for \textcolor{black}{SFDA} are applied to networks with a single prediction head. For our method with multiple prediction heads, enforcing entropy minimization for each prediction head respectively may lead to sub-optimal results when different predication heads obtain opposite results with high confidence. For example, for binary segmentation, when $h^k$ and $h^{k+1}$ predict one pixel as being the foreground with probability of 0.0 and 1.0 respectively, both branches have the lowest entropy, but their average has a high entropy. To overcome this problem, we apply entropy minimization to the mean prediction across the $K$  heads:

\begin{equation}\label{eq_lent}
    \mathcal{L}_{ment}=-\frac{1}{HW}\sum_{n=1}^{HW}\sum_{c=1}^{C}{\bar{\boldsymbol{p}}'_{c,n}}log{(\bar{\boldsymbol{p}}'_{c,n})} 
\end{equation}
where $\bar{\boldsymbol{p}}'$ is the mean probability prediction obtained by the $K$ heads  in the second forward of \textcolor{black}{TFS}. Compared with minimizing the entropy of each  prediction head respectively, minimizing the entropy of their mean prediction $\bar{\boldsymbol{p}}'$ can not only reduce the uncertainty of each head, but also encourage the consistency between them. Thus, it helps to improve the robustness of the network on samples in the target domain.

\subsection{Adaptation by Self-training}
Our adaptation method adopts a self-training process on unlabeled images in the target domain. Based on the  pseudo labels obtained by \textcolor{black}{TDG}, the overall loss function for tuning the network with \textcolor{black}{TFS} in the target domain is: 
\begin{equation}\label{eq_l_total}
    \mathcal{L} = \textcolor{black}{\mathcal{L}_{TFS}} + \lambda \mathcal{L}_{ment}
\end{equation}
where $\lambda$ is a hyper-parameter to control the weight of $\mathcal{L}_{ment}$. Note that there are two forward passes for each parameter update step, where the first forward pass obtains pseudo labels, and the loss function is calculated in the second pass for parameter update with back-propagation.  

\section{Experiments}\label{sec_exp}
\subsection{Datasets and Implementation}
\textcolor{black}{We used three datasets for experiments: 1) the public Multi-centre, multi-vendor and multi-disease cardiac image segmentation (M\&MS) dataset~\cite{campello2021multi}, where the images were acquired by devices with four different vendors, 2) an in-house Fetal Brain (FB) segmentation  dataset that contains two different MRI sequences, and 3) a public Fetal Tissue Annotation (FeTA) dataset that contains two different super-resolution methods~\cite{payette2023fetal}. A summary of these three datasets is listed in Table~\ref{tab:dataset}}. 

\subsubsection{Cardiac Image Segmentation Dataset (M\&MS)}
\textcolor{black}{The M\&MS dataset~\cite{campello2021multi} consists of 345 cardiac MRI volumes collected from six different hospitals, using four different scanner vendors, namely Siemens, Philips, General Electric, and Canon. The imaging devices were MAGNETOM Avanto for hospital 1, Achieva for hospital 2 and 3, Signa Excite, Vantage Orian, and MAGNETOM Skyra for hospital 4, 5 and 6, respectively. Following~\cite{campello2021multi}, we divide the dataset into four domains: Domain A for Siemens, comprising data from hospitals 1 and 6; Domain B for Philips, comprising data from hospitals 2 and 3; Domain C for General Electric, comprising data from hospital 4; and Domain D for Canon, comprising data from hospital 5.}
The slice number per volume varied from 10 to 13. The in-plane resolution ranged from 0.85 to 1.45 mm with slice thickness 9.2-10 mm.  ~\textcolor{black}{Following the setting in~\cite{tomar2021opttta}, we used domain A as the source domain, and B, C and D as the target domains.} The target tissues for segmentation are the Left Ventricle (LV), Right Ventricle (RV) and Myocardium (MYO). We randomly
split images in each domain into 70\%, 10\% and 20\% for training, validation and testing, respectively, and abandoned labels for the training sets in the target domains.
\begin{table*}[htbp]
  \centering
  \caption{Dice (\%) of different methods on the M\&MS dataset for cardiac structure segmentation in the target domains. The bold font highlights the best values in the first and second sections, respectively. Asterisks indicate statistical significance when comparing the methods with the Source only (*: p $\leq$ 0.05, **: p $\leq$ 0.01, ***: p $\leq$ 0.001) using a paired Student’s t-test. }
  \begin{adjustbox}{width=1.0\textwidth}
    \begin{tabular}{c|lll|lll|lll}
    \hline
    \multirow{2}[4]{*}{Method} & \multicolumn{3}{c|}{Target domain B} & \multicolumn{3}{c|}{Target domain C} & \multicolumn{3}{c}{Target domain D} \bigstrut
    \\ 
\cline{2-10} &
\multicolumn{1}{c}{LV} & \multicolumn{1}{c}{MYO} & \multicolumn{1}{c|}{RV} & \multicolumn{1}{c}{LV} & \multicolumn{1}{c}{MYO} & \multicolumn{1}{c|}{RV} & \multicolumn{1}{c}{LV} & \multicolumn{1}{c}{MYO} & \multicolumn{1}{c}{RV} \bigstrut
\\
    \hline
    \multicolumn{1}{c|}{Source only} & 87.54±10.40  & 75.50±10.97  & 81.50±16.97  & 86.47±7.61  & 77.46±7.71  & 80.55±11.44  & 88.04±6.71  & 75.88±8.57  & 77.76±17.88  \bigstrut[t] \\ 
    \multicolumn{1}{c|}{\textcolor{black}{Source only-Esb}} & \textcolor{black}{88.62±8.19} & \textcolor{black}{77.22±9.99} & \textcolor{black}{82.82±19.22} & \textcolor{black}{86.60±7.94} & \textcolor{black}{78.54±9.53} & \textcolor{black}{84.26±9.14} & \textcolor{black}{88.22±7.56} & \textcolor{black}{77.53±8.06} & \textcolor{black}{79.35±17.56} \\
    \multicolumn{1}{c|}{\textcolor{black}{Fine-tune valid}} & \textcolor{black}{90.34±6.30} & \textcolor{black}{81.68±6.53} & \textcolor{black}{85.30±12.13} & \textcolor{black}{89.54±6.06} & \textcolor{black}{82.82±4.67} & \textcolor{black}{86.20±8.20} & \textcolor{black}{88.38±8.46} & \textcolor{black}{81.08±4.10} & \textcolor{black}{83.47±11.69} \\
    \multicolumn{1}{c|}{\textcolor{black}{Fine-tune train}} & 90.90±5.36  & 83.76±5.48  & \textbf{87.63±6.11 } & \textbf{89.59±5.69}  & \textbf{83.98±4.85}  & \textbf{87.46±5.38}  & \textbf{90.68±5.40 } & \textbf{83.89±4.29 } & \textbf{85.93±5.57} 
    \\
    \multicolumn{1}{c|}{Target only} & \textbf{91.13±6.37 } & \textbf{84.37±6.56 } & 87.27±8.86  & 89.40±7.57  & 82.67±5.66  & 82.99±7.85  & 88.69±8.35  & 81.60±5.35  & 83.41±11.25 \bigstrut[b] \\
    \hline
    \multicolumn{1}{c|}{PTBN~\cite{nado2020evaluating}}  & 89.62±7.11  & 79.99±6.40**  & 82.31±15.73  & 86.06±8.96  & 79.62±7.07  & 83.76±7.34  & 88.19±6.51  & 79.03±4.92  & 81.01±11.30  \bigstrut[t]
    \\
    \multicolumn{1}{c|}{TENT~\cite{wang2021tent}}  & 89.03±8.46  & 79.96±6.44**  & 83.72±11.42  & 84.97±10.98  & 78.68±6.76  & 84.65±7.18  & 84.28±10.62  & 79.08±4.07  & 82.34±9.70 \\
    \multicolumn{1}{c|}{TTT~\cite{sun2020test}}   & 89.41±7.06  & 79.50±6.99**  & 82.72±13.27  & 85.89±9.12  & 79.57±7.10  & 83.62±6.56  & 88.13±6.93  & 79.91±4.60  & 82.31±9.63  \\
    \multicolumn{1}{c|}{URMA~\cite{fleuret2021uncertainty}}  & 90.38±5.64  & 82.09±5.39***  & 84.30±7.27  & 88.44±6.29*  & 81.73±6.21*  & 86.52±4.90*  & 88.94±5.92  & 80.69±4.74  & 83.01±7.61  \\
    \multicolumn{1}{c|}{\textcolor{black}{Ours w/o Esb}} & \textcolor{black}{90.70±5.38*} & \textcolor{black}{81.82±5.81***} & \textcolor{black}{85.73±9.22} & \textcolor{black}{89.74±3.98**} & \textcolor{black}{83.10±5.61**} & \textcolor{black}{86.69±5.50*} & \textcolor{black}{89.09±6.00} & \textcolor{black}{80.89±3.91} & \textcolor{black}{83.87±9.37} \\ 
    \multicolumn{1}{c|}{Ours } & \textbf{91.02±5.50 }* & \textbf{82.77±5.25}*** & \textbf{87.33±7.87}* & \textbf{89.64±4.03}* & \textbf{84.00±5.04}*** & \textbf{88.73±4.51}*** & \textbf{89.13±6.04} & \textbf{81.84±4.27}* & \textbf{85.30±9.53} \bigstrut[b]
    \\
    \hline
    \end{tabular}%
    \end{adjustbox}
  \label{tab:sota_dice_mms}
\end{table*}%

\subsubsection{Fetal Brain (FB) Segmentation Dataset}
The FB dataset had fetal MRI with two imaging protocols acquired from a single center, including 68 volumes of half-Fourier acquired single turbo spin-echo (HASTE) and 44 volumes of true fast imaging with steady state precession (TrueFISP).  
The slice number for each volume varied from 11 to 22, and the gestational age ranged in 21-33 weeks. The two sequences had an in-plane resolution of 0.64 to 0.70 mm and 0.67 to 1.12 mm respectively, with slice-thickness 6.5 - 7.15 mm and 6.5 mm, respectively. 
HASTE and TrueFISP were used as the source and target domains, respectively. 
We randomly split the images in each domain into 70\%, 10\% and 20\% for training, validation and testing, respectively, and abandoned the labels of training images in the target domain.

\begin{figure}
\centering
\includegraphics[width=1\linewidth]{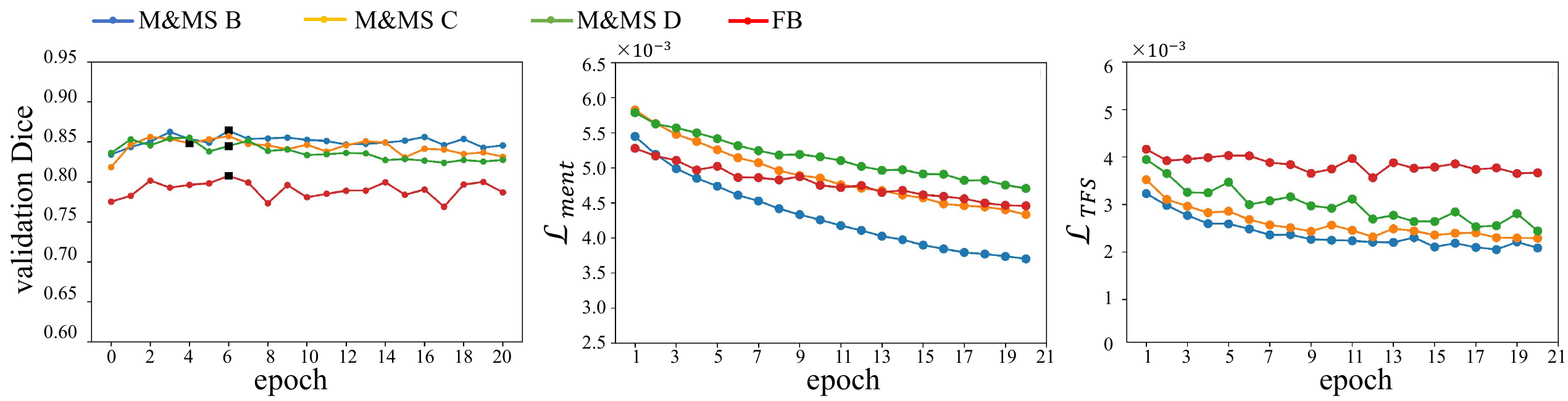}
 \begin{minipage}[t]{0.32\linewidth}\centering	(a)	 \end{minipage}
\begin{minipage}[t]{0.325\linewidth}\centering	(b)	 \end{minipage}
\begin{minipage}[t]{0.325\linewidth}\centering	(c)  \end{minipage}
\\ \vspace{0.05in}
\caption{\textcolor{black}{Evolution of validation Dice, $\mathcal{L}_{ment}$ and $\mathcal{L}_{TFS}$ during adaptation. The black squares mark the epoch with the highest validation Dice.}}	\label{fig:dice_losses_3}
\end{figure}
\subsubsection{Fetal Tissue Annotation (FeTA) Challenge Dataset}
\textcolor{black}{The FeTA Dataset~\cite{payette2023fetal} used in this study was from the FeTA2022 challenge\footnote{https://feta.grand-challenge.org/} that aims to segment seven different tissues, namely External Cerebrospinal Fluid (ECF), Grey Matter (GM), White Matter (WM), Ventricles (Ven), Cerebellum (Cer), Deep Grey Matter (DGM), and Brain Stem (BS). The official dataset has 120 samples, but only 80 samples are publicly available after the challenge, and they were acquired from the University Children's Hospital Zurich (Kispi) using 1.5T and 3T clinical GE whole-body scanners. T2-weighted single-shot Fast Spin Echo sequences were acquired with an in-plane resolution of 0.5mm $\times$ 0.5mm and a slice thickness of 3 to 5 mm. To obtain high-resolution fetal brain reconstructions, the mialSR super-resolution (SR) method~\cite{tourbier2015efficient} was used for 40 cases, while the Simple IRTK method~\cite{kuklisova2012reconstruction} was used for the other 40 cases.} \textcolor{black}{We used the 40 cases reconstructed by Simple IRTK as the source domain, and the other 40 cases reconstructed by mialSR as the target domain. For each domain, the 3D SR volumes were divided into training, validation, and testing sets in the ratio of 70\%, 10\% and 20\%, respectively.}

\begin{table*}[htbp]
  \centering
  \caption{ASSD (pixels) of different methods on the M\&MS dataset for cardiac structure segmentation in the target domains. The bold font highlights the best values in the first and second sections, respectively. Asterisks indicate statistical significance when comparing the methods with the Source only (*: p $\leq$ 0.05, **: p $\leq$ 0.01) using a paired Student’s t-test.} 
  \label{tab:sota_assd_mms}
  \begin{adjustbox}{width=1.0\textwidth}
    \begin{tabular}{c|lll|lll|lll}
    \hline
    \multirow{2}[4]{*}{Method} & \multicolumn{3}{c|}{Target domain B} & \multicolumn{3}{c|}{Target domain C} & \multicolumn{3}{c}{Target domain D} \bigstrut\\
\cline{2-10}          & \multicolumn{1}{c}{LV} & \multicolumn{1}{c}{MYO} & \multicolumn{1}{c|}{RV} & \multicolumn{1}{c}{LV} & \multicolumn{1}{c}{MYO} & \multicolumn{1}{c|}{RV} & \multicolumn{1}{c}{LV} & \multicolumn{1}{c}{MYO} & \multicolumn{1}{c}{RV} \bigstrut\\
    \hline
    \multicolumn{1}{c|}{Source only} & 0.55±0.46  & 0.64±0.45  & 0.88±1.10  & 0.58±0.37  & 0.57±0.21  & 1.18±1.38  & 0.54±0.34  & 0.59±0.30  & 1.59±2.55  \bigstrut[t]\\
    \multicolumn{1}{c|}{\textcolor{black}{Source only-Esb}} & \textcolor{black}{0.49±0.43} & \textcolor{black}{0.61±0.48} & \textcolor{black}{0.77±1.54} & \textcolor{black}{0.54±0.30} & \textcolor{black}{0.54±0.22} & \textcolor{black}{0.64±0.56} & \textcolor{black}{0.53±0.38} & \textcolor{black}{0.57±0.32} & \textcolor{black}{0.85±0.91} \\
    \multicolumn{1}{c|}{\textcolor{black}{Fine-tune valid}} & \textcolor{black}{0.43±0.36} & \textcolor{black}{0.55±0.46} & \textcolor{black}{0.49±0.44} & \textcolor{black}{0.46±0.30} & \textcolor{black}{0.49±0.18} & \textcolor{black}{0.68±0.68} & \textcolor{black}{0.53±0.43} & \textcolor{black}{0.50±0.23} & \textcolor{black}{0.69±0.60} \\
    \multicolumn{1}{c|}{\textcolor{black}{Fine-tune train}} & \textbf{0.43±0.44}  & \textbf{0.50±0.45 } & \textbf{0.43±0.29}  & \textbf{0.41±0.20}  & \textbf{0.40±0.12 } & \textbf{0.53±0.38}  & \textbf{0.36±0.16 } & \textbf{0.39±0.11 } & \textbf{0.48±0.22} \\
    \multicolumn{1}{c|}{Target only} & 0.52±0.77  & 0.55±0.61  & 0.54±0.95  & 0.40±0.24  & 0.44±0.16  & 1.31±0.99  & 0.57±0.55  & 0.51±0.24  & 0.88±0.69 
     \bigstrut[b]\\
    \hline
    \multicolumn{1}{c|}{PTBN~\cite{nado2020evaluating}}  & 0.51±0.43  & 0.60±0.46  & 0.79±1.33  & 0.65±0.44  & 0.53±0.16  & 1.03±0.92  & 0.63±0.53  & 0.55±0.26  & 1.21±1.41  \bigstrut[t]\\
    \multicolumn{1}{c|}{TENT~\cite{wang2021tent}}  & 0.61±0.69  & 0.62±0.54  & 0.67±0.99  & 0.71±0.65  & 0.59±0.24  & 0.87±0.73  & 0.88±0.72  & 0.59±0.25  & 0.55±0.31 \\
    \multicolumn{1}{c|}{TTT~\cite{sun2020test}}   & 0.48±0.38  & 0.59±0.47  & 0.82±1.23  & 0.71±0.55  & 0.55±0.19  & 1.16±0.89  & 0.64±0.53  & 0.53±0.21  & 1.01±1.20  \\
    \multicolumn{1}{c|}{URMA~\cite{fleuret2021uncertainty}}  & \textbf{0.41±0.25}* & \textbf{0.51±0.35 } & 0.50±0.23*  & 0.45±0.23*  & 0.46±0.13**  & 0.53±0.32*  & \textbf{0.47±0.22 } & 0.50±0.16  & 0.52±0.22 \\
    \multicolumn{1}{c|}{\textcolor{black}{Ours w/o Esb}} & \textcolor{black}{0.46±0.52} & \textcolor{black}{\textbf{0.58±0.57}} & \textcolor{black}{0.54±0.37*} & \textcolor{black}{0.42±0.24**} & \textcolor{black}{0.46±0.18*} & \textcolor{black}{0.62±0.43} & \textcolor{black}{0.56±0.39} & \textcolor{black}{0.51±0.21} & \textcolor{black}{0.59±0.46} \\
    \multicolumn{1}{c|}{Ours}  & 0.45±0.54  & 0.54±0.57  & \textbf{0.40±0.38}* & \textbf{0.40±0.16}** & \textbf{0.43±0.16}** & \textbf{0.39±0.26}** & 0.48±0.30  & \textbf{0.47±0.19 } & \textbf{0.46±0.29} \\
    \hline
    \end{tabular}%
    \end{adjustbox}
\end{table*}

\begin{table}[]\caption{Quantitative comparison of different SFDA methods for FB segmentation. * denotes significant difference ($p$-value $\leq$ 0.05) from Source only  using a paired Student’s t-test. }
\label{tab:sota_fb}
\centering
\begin{tabular}{
>{\columncolor[HTML]{FFFFFF}}l |
>{\columncolor[HTML]{FFFFFF}}l 
>{\columncolor[HTML]{FFFFFF}}l }
\hline
\multicolumn{1}{c|}{Methods}                          & \cellcolor[HTML]{FFFFFF}Dice (\%)   & \cellcolor[HTML]{FFFFFF}ASSD (pixel)          \\ \hline
\multicolumn{1}{c|}{Source only}                      & 84.09±6.34                         & 1.33±0.49                        \bigstrut[t]          \\
\multicolumn{1}{c|}{\textcolor{black}{Source only-Esb}} & \textcolor{black}{86.39±6.94} & \textcolor{black}{1.02±0.41} \\
\multicolumn{1}{c|}{\textcolor{black}{Fine-tune valid}} & \textcolor{black}{85.26±5.38} & \textcolor{black}{2.09±1.44} \\
\multicolumn{1}{c|}{\textcolor{black}{Fine-tune train}} & \textbf{89.71±4.87 } & \textbf{0.86±0.49} \\
\multicolumn{1}{c|}{Target only}                      & 88.85±4.12                         & 0.91±0.30                    \bigstrut[b]               \\
\hline
\multicolumn{1}{c|}{PTBN~\cite{nado2020evaluating}}                             & 85.70±4.88                          & 1.85±0.96                  \bigstrut[t]                \\
\multicolumn{1}{c|}{TENT~\cite{wang2021tent}}                             & 85.75±3.62                         & 1.60±0.71                                   \\
\multicolumn{1}{c|}{TTT~\cite{sun2020test}}                              & 85.84±4.52                         & 1.80±0.90                                    \\
\multicolumn{1}{c|}{URMA~\cite{fleuret2021uncertainty}}                             & 84.12±6.82                         & 2.18±1.19                                  \\ 
\multicolumn{1}{c|}{\textcolor{black}{Ours w/o Esb}} & \textcolor{black}{87.95±4.61} & \textcolor{black}{1.37±1.21} \\
\multicolumn{1}{c|}{Ours}                             & \textbf{89.10±3.09}*                 & \textbf{1.08±0.49} \bigstrut[b]\\ 
\hline
\end{tabular}
\end{table}
\subsubsection{Implementation Details}
All the experiments were implemented with PyTorch, 
using an NVIDIA GeForce RTX 2080Ti GPU. Our code is made available online\footnote{https://github.com/HiLab-git/UPL-SFDA}.  For M\&MS dataset and FB datasets that have a large slice thickness, we selected the widely used 2D UNet~\cite{ronneberger2015u} to demonstrate the effectiveness of our method, as most medical image segmentation models are based on UNet-like structures~\cite{litjens2017survey}. The image intensity was clipped by the 1$^{st}$ and 99-th percentiles, and linearly normalized to [-1,1]. Each slice in the M\&MS dataset was center cropped to 256$\times$256, and the slices in the FB dataset were resized to 256$\times$256. 
\textcolor{black}{For the FeTA dataset, we cropped the 3D volumes based on the brain region during preprocessing, and used the 3D U-Net architecture~\cite{cciccek20163d} for implementation. Due to memory limitation, we cropped the images to a patch size of [32, 64, 64]. In the inference stage, we applied a sliding window using the same patch size with a stride of 50\% to obtain the final segmentation results.}
During pre-training in the source domain, we trained the source model for 400 epochs with Dice loss, Adam optimizer and initial learning rate of 0.01 that was decayed to 90\% every 4 epochs. 
The model parameters with the best performance on the validation set in the source domain were used for adaptation.  
For adaptation in each target domain, we duplicated the decoder for $K$ times, and updated all the model parameters for 20 epochs with Adam optimizer and a fixed learning rate 
of $10^{-4}$. 

\textcolor{black}{The hyper-parameter setting was determined based on the labeled validation set of the target domain. Specifically,} \textcolor{black}{$K=4$ and $\lambda=1.0$. $\tau$ was set to 0.95 for the M\&MS and FeTA dataset, and 0.9 for FB dataset, respectively.  
In the adaptation stages, for M\&MS and FB dataset, we set all the slices in a single volume as a batch, and for FeTA dataset, the batch size was set to 4.}
\textcolor{black}{After training, we used the checkpoint with the best performance on the validation set for inference. Fig.~\ref{fig:dice_losses_3} shows the evolution of validation Dice, $\mathcal{L}_{ment}$ and $\mathcal{L}_{TFS}$. It can be observed that the loss functions converge in 20 epochs, and the best checkpoint was obtained at epoch 6 for M$\&$MS B, C, 4 for M$\&$MS D and 6 for the FB dataset, respectively.}

For quantitative evaluation of the volumetric segmentation results, we adopted the commonly used 
 3D Dice score and Average Symmetric Surface Distance (ASSD). As the slice thickness was large (6-10 mm) in the M\&MS and FB datasets, we calculated  ASSD values with unit of pixel. 

\begin{table*}[htbp]
  \centering
  \caption{Dice (\%) of different SFDA methods on the FeTA dataset for fetal tissue segmentation. Asterisks indicate statistical significance when comparing the methods with Source only (*: p $\leq$ 0.05, **: p $\leq$ 0.01) using a paired Student’s t-test. The bold font highlights the best values in the first and second sections, respectively.}
  \textcolor{black}{
  \begin{adjustbox}{width=1.0\textwidth}
    \begin{tabular}{c|lllllll|l}
    \hline
    Method & \multicolumn{1}{c}{ECF}   & \multicolumn{1}{c}{GM}    & \multicolumn{1}{c}{WM}    & \multicolumn{1}{c}{Ven}   & \multicolumn{1}{c}{Cer}   & \multicolumn{1}{c}{DGM}   & \multicolumn{1}{c|}{BS} & 
 \multicolumn{1}{c}{Average} \bigstrut\\
    \hline
    \multicolumn{1}{c|}{Source only} & 77.55±5.78  & 63.68±6.09  & 83.89±4.16  & 75.03±10.35  & 73.91±18.54  & 52.57±12.54  & 51.46±28.54 & 68.30±12.28 \bigstrut[t]\\
    \multicolumn{1}{c|}{Source only-Esb} & 78.03±5.65  & 64.54±6.23  & 83.89±4.19  & 76.32±8.34  & 73.88±18.68  & 47.98±13.45  & 47.95±27.99 & 67.51±12.08 \\
    \multicolumn{1}{c|}{Fine-tune valid} & 79.73±7.95  & 65.65±6.82  & 81.83±9.85  & 77.12±13.25 &  77.45±12.04  & 69.86±5.87 &  50.14±23.58 & 71.68±11.34 \\
    \multicolumn{1}{c|}{Fine-tune train} & 85.59±3.23 & 71.55±5.81  & \textbf{90.30±2.68} &  \textbf{85.18±8.03}  & \textbf{87.78±4.98}  & 81.49±6.76  & \textbf{73.17±16.58} & \textbf{82.15±6.87} \\
    \multicolumn{1}{c|}{Target only} & \textbf{86.16±2.23}  & \textbf{71.80±5.17}  & 89.93±3.11  & 83.11±9.10  & 84.38±5.37  & \textbf{82.33±5.00}  & 71.21±12.77 & 81.27±6.11 \bigstrut[b]\\
    \hline
    PTBN~\cite{nado2020evaluating}  & 77.60±7.70  & 62.54±7.73  & 82.32±5.50 &  74.06±10.63  & 80.20±12.46  & 57.91±15.46  & 52.13±23.86 & 69.53±11.91 \bigstrut[t]\\
    TENT~\cite{wang2021tent}  & 81.43±4.73*  & 65.85±5.06*  & 84.49±4.36  & 73.85±10.00 & 80.59±15.97*  & 62.26±9.12*  & 60.00±20.33 & 72.64±9.94* \\
    TTT~\cite{sun2020test}   & 80.00±5.98  & 63.77±6.53 &  83.06±4.36  & 74.29±10.39  & 81.57±12.51  & 57.57±12.81 &  56.05±22.42 & 70.90±10.71 \\
    URMA~\cite{fleuret2021uncertainty}  & 81.76±5.53* & 65.95±5.21* & \textbf{84.56±4.71 } & 73.97±9.54  & 83.02±13.10*  & \textbf{64.78±8.26}** & 62.52±19.89 & 73.79±9.46* \\
    Ours w/o Esb & 84.16±3.32  & 66.06±5.77 &  83.56±4.03  & 75.07±9.14  & 84.02±10.68  & 63.80±8.17  & \textbf{67.36±11.83} & 74.86±7.56* \\
    Ours  & \textbf{84.75±3.15}**  & \textbf{66.98±5.67}* & 83.96±3.82  & \textbf{76.66±7.61} & \textbf{84.91±8.18}* & 62.46±8.79*  & 66.57±13.52* & \textbf{75.19±7.25}** \bigstrut[b]\\
    \hline
    \end{tabular}%
    \end{adjustbox}
    }
    
  \label{tab:sota_feta_dice}%
\end{table*}%

\subsection{Comparison with State-of-the-art Methods}
To verify the effectiveness of our proposed \textcolor{black}{UPL-SFDA}, we compared it with four state-of-the-art \textcolor{black}{SFDA} methods: 1)
\textbf{PTBN}~\cite{nado2020evaluating} that updates batch normalization statistics based on unlabeled training images in the target domain without loss functions for optimization; 2)
\textbf{TENT}~\cite{wang2021tent} that only updates the parameters of batch normalization layers by minimizing the entropy of model predictions in the target domain; 3) 
\textbf{TTT}~\cite{sun2020test} that uses an auxiliary decoder to predict the rotation angle of target-domain images, and the auxiliary task's loss is used to update the model parameters; and 4)
\textbf{URMA}~\cite{fleuret2021uncertainty} that uses pseudo labels generated by a frozen main decoder to supervise auxiliary decoders. We also compared our method with \textcolor{black}{four} naive methods: 1) \textbf{Source only}  where the model pre-trained in the source domain is directly used for inference in the target domain, which serves as the lower bound; 2) \textbf{Target only} that uses training images and their labels in the target domain to train a model directly, without pre-training in the source domain; \textcolor{black}{and 3) \textbf{Fine-tune train} and 4) \textbf{Fine-tune valid} that mean the model pre-trained in the source domain is fine-tuned with the annotated training and validation sets in target domain based on fully supervised learning, respectively. 
In order to investigate the impact of ensembling, we conducted two additional experiments: 1) \textbf{Source only-Esb} that refers to ensemble based on spatial transformations of input images for inference with the pre-trained source model; 2) \textbf{Ours w/o Esb} where our method did not utilize any spatial transformations and made predictions using only one decoder.} \textcolor{black}{We implemented all the compared methods with the same backbone, i.e., UNet~\cite{ronneberger2015u} for M\&MS and FB dataset, and 3D UNet~\cite{cciccek20163d} for FeTA dataset for a fair comparison.}

\subsubsection{Result for Cardiac Image Segmentation} 
For the M\&MS dataset, we used domain A as the source domain, and adapted the pre-trained model to domain B, C and D, respectively. Table~\ref{tab:sota_dice_mms} and~\ref{tab:sota_assd_mms} show the quantitative comparison between the compared methods in terms of Dice and ASSD, respectively. It can be observed the ``Target only" outperformed ``Source only" substantially, showing the  large domain gap between the source and target domains. For example, in target domain B, ``Source only" achieved an average Dice of 87.54\%, 75.50\% and 81.50\% for LV, MYO and RV, respectively, and the corresponding Dice values obtained by ``Target only" were 91.13\%, 84.37\% and 87.27\% respectively.

\begin{figure*}[t]
	\centering
	\centerline{\includegraphics[width=17cm]{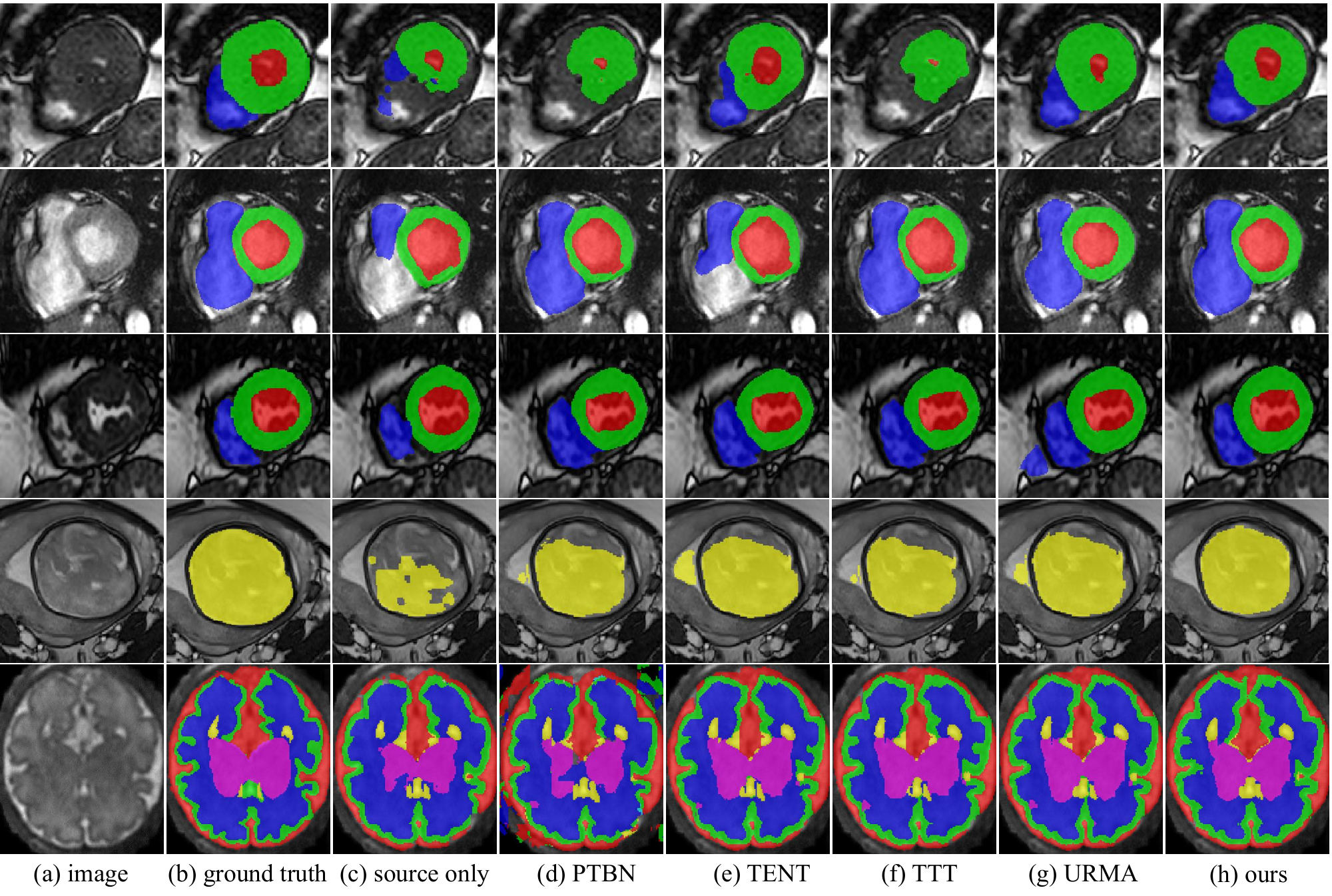}}
   \caption{Qualitative comparison of different \textcolor{black}{SFDA} methods. The top three rows are from domain B,  C and D on M\&MS dataset respectively. The last two rows are from the target domain of FB and FeTA datasets, respectively.
  } \label{fig:segmentation}
\end{figure*}
The second sections in Table~\ref{tab:sota_dice_mms} and~\ref{tab:sota_assd_mms} show that all the compared methods outperformed  ``Source only". PTBN~\cite{nado2020evaluating}, TENT~\cite{wang2021tent} and TTT~\cite{sun2020test} obtained a moderate improvement from ``Source only". For example, in Target domain B, they improved the average Dice  for LV from 87.54\% to 89.62\%, 89.03\% and 89.41\%, respectively. URMA~\cite{fleuret2021uncertainty} obtained a higher Dice (90.38\%) than these three methods, but it was inferior to our method (91.02\%).  The average Dice across the three target structures obtained by our method was  87.04\%,  87.46\% and 85.43\% in the three target domains, respectively, compared with the corresponding values of 81.51\%, 81.49\% and 80.56\%  achieved by ``source only", showing that our method improved the  average Dice scores by 5.53, 5.97 and 4.87 percentage points in the three target domains respectively.

\color{black}In terms of average Dice values, our method outperformed ``Fine-tune valid", and was close to ``Target only" ($p$-value $>$ 0.05) in target domain B, and better than  ``Fine-tune train", ``Fine-tune valid" and ``Target only" in target domain C. In target domain D, our method also outperformed ``Target only". 
Note that ``Target only", and ``Fine-tune train" require annotations in the training set of the target domain, while our adaptation method could achieve a similar  performance without the annotations.
We also analyzed the effectiveness of ensemble of multiple prediction heads with spatial transformations. Taking M\&MS B as an example, ``Source only-Esb" performed better than ``Source only", indicating the positive effect of additional data augmentations for inference. In addition, ``Ours w/o Esb" exhibited a decreased performance  compared with our complete method.  This suggests that ensembling during inference \color{black}plays a beneficial role in our approach. A visual comparison between different \textcolor{black}{SFDA} methods is shown in Fig.~\ref{fig:segmentation}. Note that ``Source only" achieved a poor performance, and the results of our method were closer to the ground truth than those of the other methods.

\subsubsection{Results for Fetal Brain Segmentation} 
\textcolor{black}{We further investigated the performance of the compared methods on FB dataset, with HASTE and TrueFISP as the source and target domains, respectively. The quantitative evaluation results are shown in Table~\ref{tab:sota_fb}. It can be observed that ``Source only" and ``Target only" achieved an average Dice of 84.09\% and 88.85\%, respectively, showing the large gap between the two domains.} ``Fine-tune train" outperformed ``Target only", achieving an average Dice of 89.71\%. The existing methods only achieved a slight improvement compared with ``Source only", with the Dice values ranging from 84.12\% to 85.84\%. In contrast, our method largely improved it to 89.10\%, which outperformed ``Target only" and  was close to ``Fine-tune train" \textcolor{black}{(p-value $>$ 0.05)}. Our method achieved an average ASSD of 1.08 pixels, which was lower than those of the other \textcolor{black}{SFDA} methods. 
The qualitative comparison in the penultimate row of Fig.~\ref{fig:segmentation} shows that the existing methods tend to achieve under-segmentation \textcolor{black}{of the fetal brain, while our method can successfully segment the entire fetal brain region with high accuracy. }
\begin{table*}[htbp]
  \centering
  \caption{Ablation study on different components of our \textcolor{black}{UPL-SFDA}. The first row (baseline) is updating the source model with pseudo labels obtained by itself and entropy minimization.
  $M$: using the binary reliability map to weight pseudo labels. \textcolor{black}{TDG: Target Domain Growing with dropout before each prediction head}. $\mathcal{T}$: Random spacial transformation for each prediction head.}
  \begin{adjustbox}{width=1.0\textwidth}
    \begin{tabular}{ccccc|ccc|c|ccc|c}
    \hline
    \multicolumn{5}{c|}{Components}       & \multicolumn{4}{c|}{Dice (\%)} & \multicolumn{4}{c}{ASSD (pixel)} \bigstrut\\
    \hline
    M     & \textcolor{black}{TDG}   & $\mathcal{T}$     & \textcolor{black}{TFS}   & $L_{ment}$ & M\&MS B & M\&MS C &  M\&MS D & FB    &  M\&MS B & M\&MS C & M\&MS D & FB \bigstrut\\
    \hline
          &       &       &       &       & 84.10±9.67 & 84.20±6.16 & 83.37±6.90 & 83.44±7.38 & 0.60±0.84 & 0.64±0.36 & 0.60±0.38 & 1.39±0.76 \bigstrut[t]\\
    \checkmark    &       &       &       &       & 84.49±10.35 & 85.09±5.79 & 84.10±6.97 & 85.88±5.49 & 0.58±0.80 & 0.49±0.21 & 0.56±0.36 & 1.56±1.01 \\
    \checkmark    & \checkmark    &       &       &       & 84.52±10.31 & 85.22±5.57 & 84.12±6.89 & 86.77±4.17 & 0.56±0.63 & 0.50±0.22 & 0.55±0.34 & 0.95±0.24 \\
    \checkmark    & \checkmark    & \checkmark    &       &       & 86.39±7.90 & 86.16±6.17 & 84.98±7.08 & 86.92±5.41 & \textbf{0.44±0.43} & 0.46±0.25 & 0.48±0.26 & 0.91±0.36 \\
    \textcolor{black}{\checkmark} & \textcolor{black}{\checkmark} & \textcolor{black}{\checkmark} & \textcolor{black}{\checkmark} & & \textcolor{black}{86.52±6.63} & \textcolor{black}{86.79±4.44} & \textcolor{black}{85.20±7.02} & \textcolor{black}{88.11±5.06} & \textcolor{black}{0.46±0.44} & \textcolor{black}{0.42±0.17} & \textcolor{black}{0.48±0.26} & \textcolor{black}{\textbf{0.86±0.34}} \\
    \checkmark    & \checkmark    & \checkmark    & \checkmark    & \checkmark    & \textbf{87.04±6.20} & \textbf{87.46±4.52} & \textbf{85.43±6.61} & \textbf{89.10±3.09} & 0.46±0.38 & \textbf{0.40±0.24} & \textbf{0.47±0.33} & 1.08±0.49 \\
    

    \textcolor{black}{\checkmark} & \textcolor{black}{\checkmark} & & \textcolor{black}{\checkmark} & \textcolor{black}{\checkmark} & \textcolor{black}{85.16±6.91} & \textcolor{black}{85.43±6.54} & \textcolor{black}{84.42±6.89} & \textcolor{black}{87.57±3.20} & \textcolor{black}{0.49±0.32} & \textcolor{black}{0.56±0.37} & \textcolor{black}{0.54±0.36} & \textcolor{black}{1.02±0.27} \\
    \textcolor{black}{}      & \textcolor{black}{\checkmark} & \textcolor{black}{\checkmark} & & \textcolor{black}{\checkmark} & \textcolor{black}{85.82±7.95} & \textcolor{black}{86.94±5.13} & \textcolor{black}{84.09±7.87} & \textcolor{black}{84.48±6.88} & \textcolor{black}{0.46±0.42} & \textcolor{black}{0.41±0.20} & \textcolor{black}{0.55±0.39} & \textcolor{black}{1.52±0.92} \bigstrut[b]\\

    \hline
    \multicolumn{5}{c|}{Source only}      & 81.51±12.78 & 81.47±8.92 & 80.56±11.05 & 84.09±6.34 & 0.69±0.67 & 0.71±0.65 & 0.90±1.06 & 1.33±049 \bigstrut[t]\\
    \multicolumn{5}{c|}{Target only}      & 87.59±7.26 & 85.02±7.02 & 84.56±8.31 & 88.85±4.12 & 0.53±0.77 & 0.71±0.46 & 0.65±0.49 & 0.91±0.30 \bigstrut[b]\\
    \hline
    \end{tabular}%
    \end{adjustbox}
  \label{tab:ablation}%
\end{table*}%

\subsubsection{Results for 3D Fetal Tissue Segmentation} 
Quantitative evaluation results on the FeTA dataset in terms of Dice are shown in Table~\ref{tab:sota_feta_dice}. It shows that ``Source only" and ``Target only" achieved an average Dice of 68.30\% and 81.27\%, respectively, indicating the large gap between the two domains. 
Our method increased the average Dice by 6.89 percentage points compared with ``Source only", reaching 75.19\%. In contrast, the existing methods had a lower performance than ours. The average Dice obtained by PTBN~\cite{nado2020evaluating}, TENT~\cite{wang2021tent}, TTT~\cite{sun2020test} and URMA~\cite{fleuret2021uncertainty} was 69.53\%, 72.64\%, 70.90\% and  73.79\%, respectively.
 The qualitative comparison in the last row of Fig.~\ref{fig:segmentation} demonstrates that our method outperformed the other methods in terms of segmentation performance.

\begin{figure}
	\centering
\includegraphics[width=0.99\linewidth]{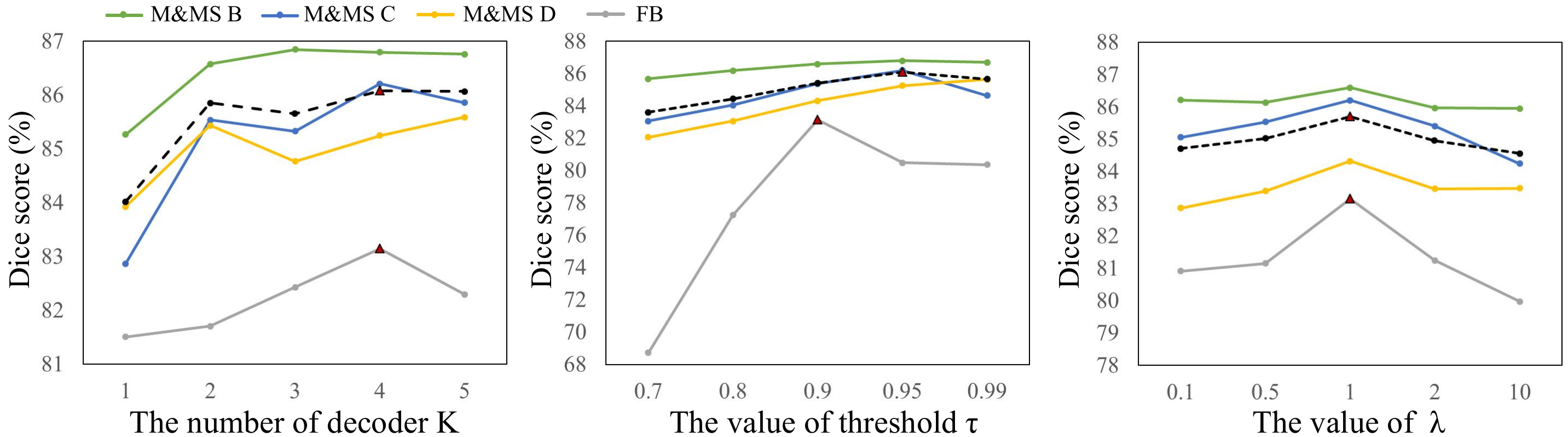}
	\begin{minipage}[t]{0.32\linewidth}\centering	(a)	 \end{minipage}
    \begin{minipage}[t]{0.32\linewidth}\centering	(b)	 \end{minipage}
	\begin{minipage}[t]{0.32\linewidth}\centering	\textcolor{black}{(c)}  \end{minipage}
	\\ \vspace{0.05in}
	\caption{Performance of our method with different hyper-parameter values on the validation sets of different target domains.	}
	\label{fig:selection_of_k_t}
\end{figure} 
\subsection{Ablation Analysis of Our UPL-SFDA}


\begin{figure*}[t]
	\centering
	\centerline{\includegraphics[width=18cm]{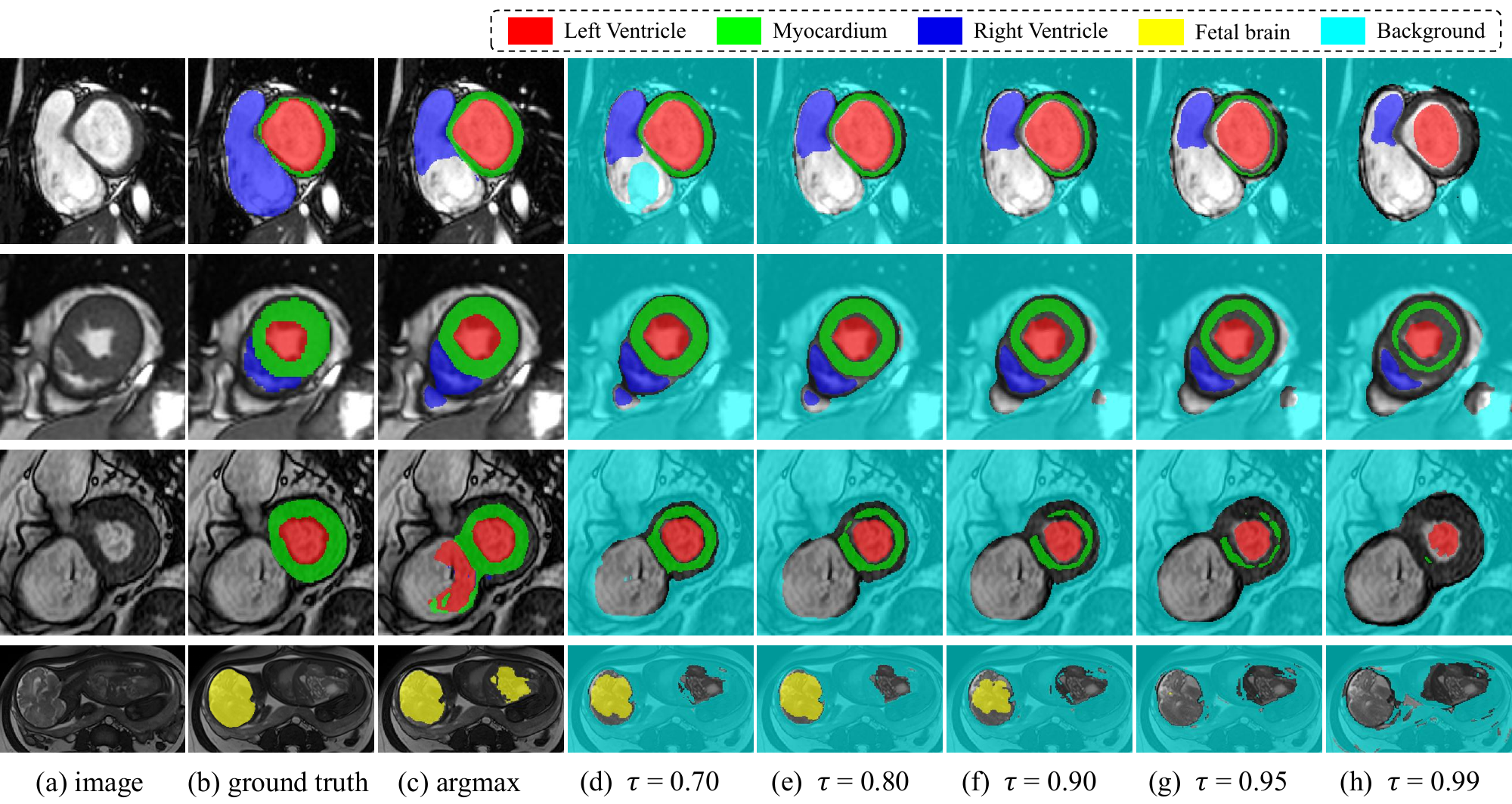}}
	\caption{Effect of confidence threshold $\tau$ on reliable pseudo labels. The first three rows are from domain B, C and D on M\&MS dataset respectively, and the last row is from the target domain of FB dataset. (c) shows pseudo labels obtained by argmax, and (d)-(h) are reliable pseudo labels with different $\tau$ values, where uncolored regions are pixels with unreliable pseudo labels.} \label{fig:pseudo label on thr}
\end{figure*}

\subsubsection{Effect of Hyper-parameters}
\textcolor{black}{There are three important hyper-parameters specific to our method: the number of duplicated prediction heads $K$, the confidence threshold $\tau$ to select reliable pseudo labels for supervision, and the loss weight $\lambda$.}
We first investigated the effect of $K$ by setting it to 1 to 5 respectively, and the performance on the validation sets of the two datasets are shown in Fig.~\ref{fig:selection_of_k_t}(a). It can be observed that $K=1$ performed worse than larger $K$ values, showing the superiority of using TDG. The performance on both datasets improved when $K$ changed from 1 to 4, and $K=5$ did not further bring  performance improvement. Therefore, we set $K=4$ for our method.

Then we investigated how $\tau$ affected the pseudo labels and the \textcolor{black}{SFDA} performance. 
Fig.~\ref{fig:pseudo label on thr} shows some examples of reliable pseudo labels with different $\tau$ values. We found that a higher threshold $\tau$ will lead to smaller reliable regions for each class, which helps to avoid the model being affected by unreliable regions of the pseudo labels. Quantitative comparison between different $\tau$ values is demonstrated in Fig.~\ref{fig:selection_of_k_t}(b), which shows that the performance on the M\&MS dataset was relatively stable with different $\tau$ values, and $\tau=0.95$ performed slightly better than the other values in average. The best $\tau$ value on the FB dataset was 0.90 based on performance on the validation set. Therefore, we set $\tau$ to 0.95 and 0.9 for the two datasets, respectively. \textcolor{black}{The performance on the validation set with different $\lambda$ values is shown in Fig.~\ref{fig:selection_of_k_t}(c). It demonstrates that the best $\lambda$ was 1.0 for the different datasets.}



\begin{figure*}[t]
	\centering
	\centerline{\includegraphics[width=17cm]{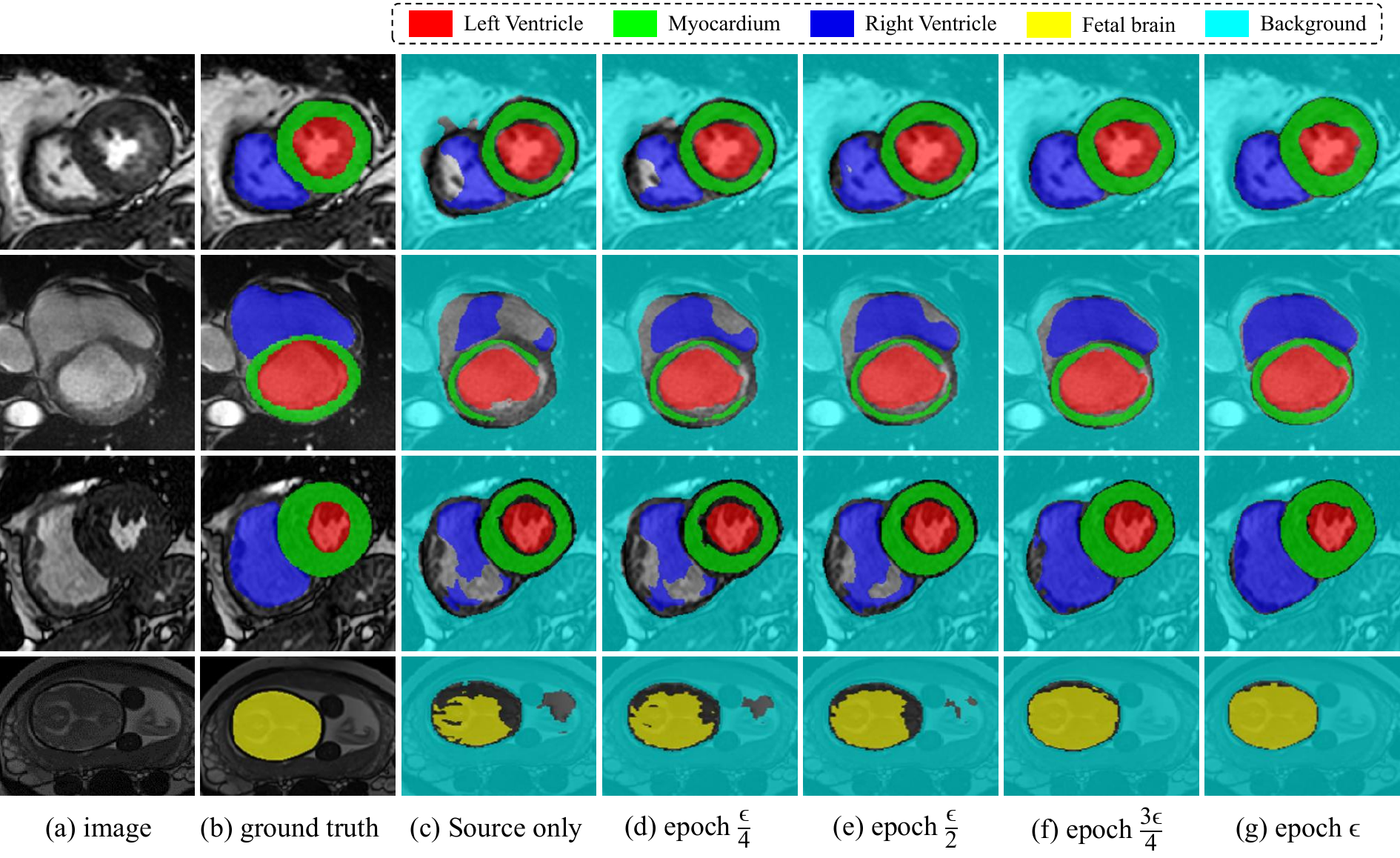}}
	\caption{Pseudo labels at different training steps in self-training.  
    $\epsilon$ means the epoch number with the highest performance on the validation set. 
    The first three rows are from domain B, C and D of M\&MS dataset respectively, and the bottom row is from the target domain of FB dataset. 
    In (c)-(g), only reliable pseudo labels are encoded by colors, and pixels without encoded colors will be ignored in the calculation of \textcolor{black}{TFS} loss. 
 } \label{fig:self training}
\end{figure*}

Fig.~\ref{fig:self training} shows the reliable pseudo labels obtained at different training epochs in the target domains. It can be observed that
the pseudo labels are  updated during the self-training process, and their quality  gradually improves at different training epochs. In addition, the confidence of the pseudo labels also improves with the increase of training epochs.

\subsubsection{Ablation study of each component}
To evaluate the effectiveness of each of the proposed components in our \textcolor{black}{UPL-SFDA}, we set the baseline as updating the source model based on self-training where the network was supervised by its own prediction and an entropy minimization loss. 
The quantitative results obtained by different variants 
of our method are shown in Table~\ref{tab:ablation}, where
$M$ means using the binary reliability map to weight pseudo labels, \textcolor{black}{TDG} means using target domain growing with dropout before each prediction head, and $\mathcal{T}$ means using random spatial transformation for each prediction head. $\mathcal{L}_{ment}$ means minimizing entropy of the mean prediction across the $K$ heads, rather than minimizing  entropy of each head respectively. 

Table~\ref{tab:ablation} shows that each component of our method led to a performance improvement. Take the performance on the domain C of M\&MS dataset as an example, the average Dice score obtained by ``Source only" was 81.47\%. 
\textcolor{black}{The baseline obtained an average Dice of 84.20\%, and introducing reliability map weighting for pseudo labels improved it to 85.09\%. For TDG, only using dropout for perturbations obtained an average Dice of 85.22\%, and additionally using spatial transformation for the prediction heads improved it to 86.16\%, showing that the spatial transformation plays an important role in our method. Then, using our Twice Forward} \textcolor{black}{ pass Supervision (TFS) loss improved it to 86.79\%, and our proposed method combining all these modules with $\mathcal{L}_{ment}$ obtained the highest Dice score of 87.46\%. 
Note that by removing the spatial transformation for the prediction heads in our method, the average Dice decreased to 85.43\%.}
\textcolor{black}{We also tried to only combine $L_{ment}$ loss with TDG using the spatial transformations (i.e., removing TFS loss), and the average Dice dropped to 86.94\%. } In addition, Table~\ref{tab:ablation} shows that our method outperformed ``Target only" on domains C and D in the M\&MS dataset and the target domain of FB dataset in terms of average Dice score. 

\section{Discussion}\label{sec:discussion}

Our proposed \textcolor{black}{UPL-SFDA} based on \textcolor{black}{TDG} and reliable pseudo label supervision has several advantages over existing \textcolor{black}{SFDA} methods for domain adaptation without access to source-domain images. First, unlike some existing methods~\cite{sun2020test,karani2021test,he2021autoencoder} using auxiliary branches in the network that require special training strategies in the source domain, our method does not require training auxiliary branches in the source domain, and it makes the training methods in the source and target domains independent. This decoupling makes our method more general to a wider range of pre-trained models. Second, compared with existing methods using entropy minimization for regularization~\cite{wang2021tent}, our method uses reliable pseudo labels for adaptation, which provides more effective supervision signals for model update. In addition, based on the \textcolor{black}{TDG} strategy with perturbations, we obtain multiple predictions that can provide high-quality pseudo labels with efficient uncertainty estimation, which prevents the model being corrupted by unreliable pseudo labels. Using entropy minimization on the average prediction across the multiple heads can encourage a consistency between them, which also improves the robustness of our method.


\color{black}{The pseudo label-based supervision loss $\mathcal{L}_{w-dice}$ and the unsupervised regularization loss $\mathcal{L}_{ment}$ have two similarities. First, both of them are based on multi-head agreement. $\mathcal{L}_{w-dice}$ uses relatively consensus regions of the $K$ prediction heads as pseudo labels, and  $\mathcal{L}_{ment}$ encourages the $K$ prediction heads to obtain consensus results by minimizing the uncertainty in the average prediction. Second, the two terms will increase the confidence of the predictions. $\mathcal{L}_{w-dice}$ drives the predictions to be closer to the hard pseudo labels, while $\mathcal{L}_{ment}$ directly minimizes the entropy, and both of them will reduce uncertain predictions. However, they also have several important differences. First, $\mathcal{L}_{w-dice}$ encourages consistency across two different forward passes with feature perturbations, while   $\mathcal{L}_{ment}$  is for consistency across prediction heads. Second, $\mathcal{L}_{w-dice}$ is applied to high-confidence pixels (with a threshold of $\tau$), while  $\mathcal{L}_{ment}$ is applied to the entire image region. Thirdly, $\mathcal{L}_{w-dice}$ is a pseudo label-based supervision loss, while $\mathcal{L}_{ment}$ is an unsupervised loss for regularization. Therefore, the two terms are complementary to each other.}

\color{black}Introducing perturbations to the $K$ prediction heads in TDG is important for achieving good performance. Without perturbation, the $K$ prediction heads will obtain the same result, which degrades to  just using the pre-trained model with a single prediction head. With  perturbations, the $K$ prediction results are different and their ensemble is more robust, which can overcome the bias in each prediction head and lead to uncertainty estimation. 
In addition, we implemented our \textcolor{black}{TDG} with an encoder-decoder structure due to that most state-of-the-art CNNs for medical image segmentation have an encoder-decoder structure~\cite{ronneberger2015u,Fabian2021}. It may also be applied to other networks~\cite{Xie2021cotr} by duplicating the prediction head multiple times with perturbations in the target domain. 

\textcolor{black}{In our experiment, a validation set  with annotations in the target domain is used to select hyper-parameters for the compared methods. The advantage of using the labeled validation set is that it allows to find the optimal hyper-parameters such as learning rate and weights of loss terms of each compared method. In addition, it allows early stopping and checkpoint selection to avoid over-fitting on the training set in the target domain, which ensures a fair comparison between the different methods. One may also use the validation set to update the model weights by fine-tuning, which could provide more supervision signal directly to the model for parameter optimization. However, it may lead the model to over-fit the validation set that is usually small. In addition, using the validation set for hyper-parameter selection rather than model learning is a work standard in the machine learning community. However, in some cases, the labeled validation set may not be available, making it less practical to use the validation set to fine-tune the pre-trained model.}



\textcolor{black}{This work still has some limitations. First, our method involves performing two forward passes for each gradient back-propagation, which takes more time than using a single forward pass. The training time consumption for our method is }\textcolor{black}{slightly higher than TENT~\cite{wang2021tent}, but lower than URMA~\cite{fleuret2021uncertainty}. For instance, in M\&MS B, our method takes an average of 0.661s per case to train one epoch, while TENT and URMA require 0.342s and 0.944s in average, respectively. The average inference time for our method is 0.342s per case, and slightly higher than TENT's 0.269s. Second, we have employed a labeled validation set in the target domain to select the optimal hyper-parameters. However, in practical applications, acquiring a validation set could be challenging, making it hard to determine hyper-parameters. Additionally, TDG with multiple prediction heads increase the memory cost, which does not allow a large patch size or batch size for dealing with 3D medical images and may limit the performance.}
\section{Conclusion}\label{sec:conclusion}
\color{black}In conclusion, we propose a novel uncertainty-aware pseudo label-guided approach for \textcolor{black}{Source-Free Domain Adaptation (UPL-SFDA)} in medical image segmentation, which uses target domain growing to generate multiple predictions for an input to obtain reliable pseudo labels with a weight map based on uncertainty estimation. The network is supervised by the weighted pseudo labels and minimizing the entropy of the average of the multiple predictions. A twice forward pass supervision strategy is also proposed to avoid the network being biased towards its own predictions in self-training. Experimental results on multi-site heart MRI segmentation and cross-modality fetal brain segmentation showed that our method outperformed existing \textcolor{black}{SFDA} methods, and it was comparable to and even better than supervised training in the target domain. In the future, it is of interest to apply our method to other segmentation tasks. 

\bibliographystyle{IEEEtran}
\bibliography{myref}
\end{document}